# Spiking neural networks with Hebbian plasticity for unsupervised representation learning

Naresh Ravichandran[1], Anders Lansner[1,2], Pawel Herman[1,3] *

1- Division of Computational Science and Technology, School of Electrical Engineering and Computer Science, KTH Royal Institute of Technology, Sweden
2- Department of Mathematics - Stockholm University, Sweden
3- Digital Futures, KTH Royal Institute of Technology, Sweden

**Abstract**. We introduce a novel spiking neural network model for learning distributed internal representations from data in an unsupervised procedure. We achieved this by transforming the non-spiking feedforward Bayesian Confidence Propagation Neural Network (BCPNN) model, employing an online correlation-based Hebbian-Bayesian learning and rewiring mechanism, shown previously to perform representation learning, into a spiking neural network with Poisson statistics and low firing rate comparable to *in vivo* cortical pyramidal neurons. We evaluated the representations learned by our spiking model using a linear classifier and show performance close to the non-spiking BCPNN, and competitive with other Hebbian-based spiking networks when trained on MNIST and F-MNIST machine learning benchmarks.

## 1  Introduction

The success of deep learning (DL) in solving various real-world pattern recognition benchmarks has shown the importance of building large-scale artificial neural networks (ANNs) with the ability to learn distributed internal representations from real-world data. One of the emerging concerns however is the energy footprint of heavy computations involved in training large ANN architectures. In response to this challenge there has been growing interest in neuromorphic approaches that build on more biologically plausible spiking neural networks (SNNs). This new generation of neural network models holds a promise for energy-efficient neuromorphic hardware that can handle real-time data streams efficiently with sparse and asynchronous event-based communication [1]. It is therefore imperative that, in parallel to DL development, we develop SNNs that can learn representations from real-world data. Building such SNN models has been typically addressed either by converting a traditional non-spiking ANN trained with gradient descent learning into a SNN, or by modifying backprop-based gradient descent algorithms to accommodate spiking neurons [1,2]. Since the

* This work is funded by the Swedish e-Science Research Centre (SeRC), Vetenskapsrådet (Swedish Research Council) Grants No. 2018-05360 and 2018-07079 (Indo-Swedish joint network grant 2018). Simulations were performed at the PDC Center for High Performance Computing, KTH Royal Institute of Technology and at Vega at the Institute of Information Science (Slovenia). We would like to thank Jing Gong, Artem Zhmurov, and Lilit Axner (ENCCS) and Martin Rehn (KTH) for collaborating and developing the code with GPU acceleration.

current approaches do not fully leverage the biological nature of the learning principles in SNNs, there is a motivated concern that full potential of SNNs and their neuromorphic implementations may not be harnessed.

Our philosophy for SNN design is steeped into the biological brain's inspirations and hence we aim to develop a biologically constrained SNN model that performs unsupervised representation learning based on Hebbian learning principles. For this, we derive our model from an abstract (non-spiking) brain-like BCPNN architecture, previously shown to perform representation learning by solely using Hebbian learning (synaptic plasticity) and Hebbian rewiring (structural plasticity) mechanisms [5]. Crucially, we employ on-line Hebbian learning directly in the spiking domain. To this end, we interpret spikes as stochastic independent samples from a Poisson distribution, where the underlying firing rates are computed as probabilities from the BCPNN model. This is motivated from the observations that *in vivo* cortical pyramidal neurons show reliable firing rate whereas the timing of spikes is highly irregular and the corresponding inter-spike intervals closely resembles a Poisson distribution [3,4]. Our main contribution is to show that the BCPNN model can be converted to a SNN preserving the biological details with minimal compromise on performance. The spiking statistics in our model reach a maximum firing rate of around 50 spikes/s, matching the sparse firing of *in vivo* cortical pyramidal neurons. We evaluated the internal representation of the model by means of a linear classifier and compared it with the corresponding non-spiking model as well as other SNNs with Hebbian learning.

## 2   Model description

We summarize key details of the model relevant to the spiking version and refer to previous works on the feedforward non-spiking BCPNN model for full details [5].

**Modular network design**: Our spiking BCPNN model consists of one spiking input layer and one spiking hidden layer. The layer architecture is derived from the columnar organization of the neocortex. Each layer in our network model is composed of many identical hypercolumns modules, each of which in turn comprises many neuron units (referred to as minicolumns) sharing the same receptive field.

**Localized learning**: The learning mechanism is local, online, and correlation-based Hebbian-Bayesian synaptic plasticity where each synapse accumulates short and long-term traces of pre-, post-, and joint activities. From the pre- and post-synaptic spikes at time $t$, $S_i, S_j \in \{0, 1\}$, we compute $Z$-traces, $Z_i$ and $Z_j$, as a form of short-term filtering ($\tau_z$ ~ few milliseconds) providing a coincidence detection window between pre- and post-synaptic spikes for subsequent LTP/LTD induction (Eq. 1). The $Z$-traces are further transformed into $P$-traces, $P_i, P_j$, and $P_{ij}$, with long time-constants ($\tau_p$ ~ seconds to hours) reflecting LTP/LTD synaptic processes (Eq. 2). The $P$-traces are finally transformed to bias and weight parameter of the synapse corresponding to terms in

ANNs (Eq. 3). All the spike and trace variables are time dependent (time index is dropped for the notation brevity).

$$\tau_z \frac{dZ_i}{dt} = \frac{\tau_z}{\Delta t} S_i - Z_i, \qquad \tau_z \frac{dZ_j}{dt} = \frac{\tau_z}{\Delta t} S_j - Z_j, \qquad (1)$$

$$\tau_p \frac{dP_i}{dt} = Z_i - P_i, \qquad \tau_p \frac{dP_{ij}}{dt} = Z_i Z_j - P_{ij}, \qquad \tau_p \frac{dP_j}{dt} = Z_j - P_j, \qquad (2)$$

$$b_j = \log P_j, \qquad w_{ij} = \log \frac{P_{ij}}{P_i P_j}, \qquad (3)$$

**Localized rewiring:** The synaptic rewiring mechanism adaptively finds efficient sparse connectivity between the layers, mimicking structural plasticity in the brain [5]. This mechanism uses the $P$-traces locally available at each synapse to maximize a "usage" score and updates the sparse binary connectivity matrix $c_{ij}$ accordingly.

**Neuronal activation:** The total input $I_j$ for neuron $j$ is updated to be weighted sum of incoming spikes with the time-constant $\tau_z$ (acting here as the depolarization time constant) (Eq. 4). The activation of the neuron, $\pi_j$, is computed as a softmax function of the input $I_j$ (Eq. 5), which induces a soft-winner-takes-all competition between the minicolumn units within each hypercolumn module. The output of the softmax function reflects the posterior belief probability of the minicolumn unit according to the BCPNN formalism [5]. In the non-spiking (rate-based) BCPNN model, this activation $\pi_j$ acts as the firing rate and can be directly communicated as the neuronal signal. For SNNs, we independently sample binary values from this $\pi_j$ activation probability scaled by the maximum firing rate $f_{max}$ for each time step (Eq. 6). Note that when $f_{max} = 1000$ spikes/s (and $\Delta t = 1$ms), the spike generation process from Eq. 6 is simply a stochastic sampling of the underlying firing rate probability and setting $f_{max} < 1/\Delta t$ linearly scales the activation probability to the maximum firing rate. Also, in both learning (Eq. 1) and synaptic integration (Eq. 4) steps, we scaled the binary spiking signal by $\tau_z/\Delta t$ as this renders the filtered spike statistics of model to be equivalent to the rate model.

$$\tau_z \frac{dI_j}{dt} = b_j + \frac{\tau_z}{\Delta t} \sum_{i=0}^{N_i} S_i\, w_{ij}\, c_{ij} - I_j, \qquad (4)$$

$$\pi_j = \frac{\exp(I_j)}{\sum_{k=1}^{Mh} \exp(I_k)}, \qquad (5)$$

$$S_j \sim P(\text{spike between t and t} + \Delta t) = \pi_j\, f_{max}\, \Delta t \qquad (6)$$

## 3 Experiments

### 3.1 Comparison of classification performance

To benchmark our spiking BCPNN model on the MNIST (hand-written digit images) and F-MNIST (fashion apparel images) datasets, we first trained it in a purely unsupervised manner (representation learning) and then used a linear classifier (cross entropy loss, SGD with Adam optimizer, 100 training epochs) to predict class labels ($n$ = 3 randomized runs, all parameters are listed in Table 1). Table 2 shows that the classification accuracy of our model is competitive with the non-spiking BCPNN as well as other SNNs with Hebbian-like plasticity (STDP and its variants).

| Type | Parameter | Value | Description |
|---|---|---|---|
| Synaptic | $\tau_z$ | 20 ms | Short-term filtering time constant |
| | $\tau_p$ | 5 s | Long-term learning time constant |
| | $p_{conn}$ | 10 % | Sparse connectivity between layers |
| Neuronal | $H_i, M_i$ | 784, 2 | N:o input layer hypercolumns & minicolumns |
| | $H_h, M_h$ | 100, 100 | N:o hidden layer hypercolumns & minicolumns |
| | $f_{max}$ | 50 spikes/s | Maximum firing rate |
| Training protocol | $\Delta t$ | 1 ms | Simulation time step |
| | $T_{pat}$ | 200 ms | Time period for each pattern |
| | $T_{gap}$ | 100 ms | Time period of gap between patterns |
| | $N_{epoch}$ | 10 | N:o of training epochs |
| | $N_{pat}$ | 60000 | N:o of training patterns |

Table 1: Network parameters.

| Model | Activity | Plasticity | MNIST | F-MNIST |
|---|---|---|---|---|
| BCPNN (this work) | spiking | BCPNN | 97.7 ± 0.09 | 83.8 ± 0.12 |
| BCPNN | rate | BCPNN | 98.6 ± 0.08 | 89.9 ± 0.09 |
| Diehl & Cook, 2015 [6] | spiking | STDP | 95.0 | -- |
| Kheradpisheh et al., 2018 [7] | spiking | STDP | 98.4 | -- |
| Mozafari et al., 2019 [8] | spiking | STDP-R | 97.2 | -- |
| Hao et al., 2019 [9] | spiking | sym-STDP | 96.7 | 85.3 |
| Dong et al., 2023 [10] | spiking | STB-STDP | 97.9 | 87.0 |

Table 2: Linear classification accuracy (%).

### 3.2 Spiking BCPNN with sparse firing learns distributed representations

In Fig. 1A we plotted the neuronal support, i.e., input, $I_j$, superimposed with the spiking output, $S_j$, for 30 randomly selected neurons within a single hypercolumn module after training the network on MNIST data (for visualization, we offset each neuron's input by 50mV vertically, scaled them to be in the range -80mV to -55mV and added a spike event of 40 mV) and observed sparse spiking with occasional bursts. In Fig. 1B we plotted the firing rate of each neuron in a single randomly chosen hypercolumn module by convolving the spike train with a Gaussian kernel ($\sigma$ = 50ms). We see that most neurons have low firing rates (~2 spikes/s), with a very few (typically one) neurons showing high level of activity (~50 spikes/s) within the duration of a stimulus pattern presentation (gray vertical bars) due to the local competition within the hypercolumn. We plotted the receptive fields for three hypercolumns and the filters learned by six minicolumns each (randomly chosen) in Fig. 1C. They provide a good qualitative match to the previously published results of the non-spiking BCPNN model [5].

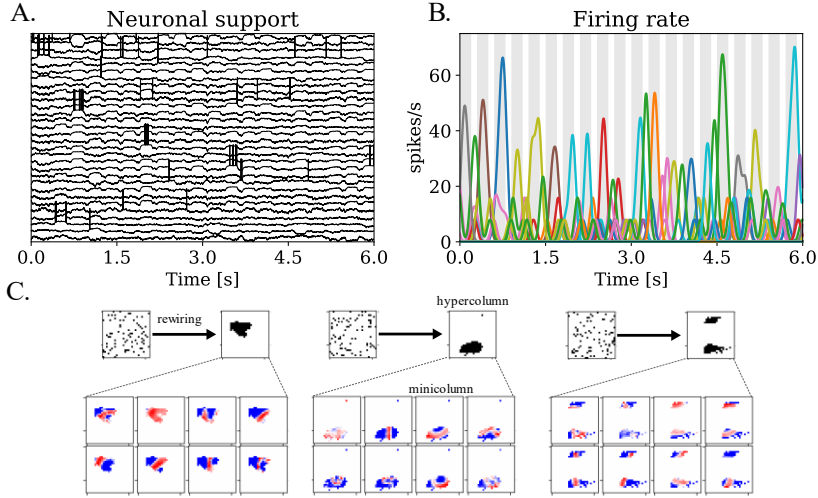

**Figure 1 A.** Neuronal support recorded after training for a time period of 6s across 30 randomly selected neurons shows sparse spiking activity. **B.** Firing rate computed from all ($M_h$=100) neurons within one hypercolumn. For the duration of pattern presentations (gray vertical bars, $T_{pat}$ = 200ms), mostly a single neuron shows a high firing rate while the rest are at a baseline firing rate. **C.** Local receptive fields are formed from randomly initialized connections through the rewiring mechanism, and individual minicolumns learn filters within their receptive field resembling orientation edge detectors.

### 3.3 Filtering enables spiking models to approximate the non-spiking model

We studied the effects of short-term filtering (Z-traces) in terms of classification performance (Fig. 2). We ran our experiments by training on a reduced version of MNIST dataset with 1000 training and 1000 test patterns while varying $\tau_z$ and $f_{max}$ (all other parameters same as in Table 1). For biologically realistic values of $f_{max}$, like 50 spikes/s, performance with $\tau_z \leq$ 10ms is very low ($\tau_z$ = 1ms is effectively no filtering). This is because pre- and post- synaptic spikes are expected to coincide within this time-window for learning to occur, whereas spikes are generated sparsely and irregularly from a Poisson distribution. However, for $\tau_z \sim$50ms, the performance closely approximates the non-spiking model since this time window is sufficient to expect pre- and post-synaptic spikes to coincide and be associated. For $f_{max} >$ 500Hz (non-biological case), accuracy is high for $\tau_z$ over a wider range since the spikes are dense samples of the underlying neuronal activation and short-term filtering is not necessarily helpful. All models irrespective of $f_{max}$ drop sharply in performance after $\tau_z >$ 100ms, very likely because the temporal window provided is too long compared to the presentation time of each pattern ($T_{pat} + T_{gap}$ = 300ms) and the learning wrongly associates spikes of a pattern with spikes from previous patterns.

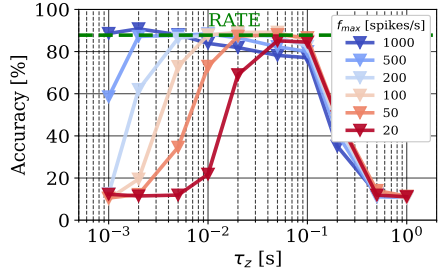

**Figure 2:** Effect of short-term filtering on classification performance.

## 4  Conclusion

We have demonstrated that our spiking BCPNN model can learn internal representations, preserving the learning and rewiring mechanisms introduced in the non-spiking BCPNN model, offering competitive classification performance. Our Poisson spike generation mechanism is simpler than integrate-and-fire models, but it still recapitulates *in vivo* irregular cortical pyramidal spiking patterns with realistic firing rates. We suggest that it is the Hebbian plasticity mechanism that provides a robust learning algorithm tolerating the highly irregular sparse spiking. This is in stark contrast to backprop-based algorithms where it is not straightforward to accommodate spiking neurons. We found that short-term filtering (*Z*-traces) was crucial for this process. The time constants we found to work best ($\tau_z$ ~50ms) roughly match the dendritic depolarization time constant (paralleling the integration step in Eq. 4), and the NMDA-dependent $Ca^{2+}$ influx required for synaptic plasticity (learning step in Eq. 1).

Our scaling experiments (not shown) suggested that the network scales well in terms of performance although the running time is 100x slower compared to the non-spiking model since the timestep needs to be around 1ms (simulations took ~10 hours on custom CUDA code running on A100 GPUs). More efficient software and custom hardware implementation can make large-scale SNN simulations more efficient. Another direction of interest is in developing a more complex network architecture that combines recurrent attractors implementing associative memory with hierarchical representation learning (unsupervised) networks.

## References

[1] Roy, Kaushik, Akhilesh Jaiswal, and Priyadarshini Panda. "Towards spike-based machine intelligence with neuromorphic computing." *Nature* 575.7784 (2019): 607-617.

[2] Tavanaei, Amirhossein, et al. "Deep learning in spiking neural networks." *Neural networks* 111 (2019): 47-63.

[3] Shadlen, Michael N., and William T. Newsome. "The variable discharge of cortical neurons: implications for connectivity, computation, and information coding." *Journal of neuroscience* 18.10 (1998): 3870-3896.

[4] Churchland, Mark M., et al. "Stimulus onset quenches neural variability: a widespread cortical phenomenon." *Nature neuroscience* 13.3 (2010): 369-378.

[5] Ravichandran, N. B., Lansner, A., & Herman, P. (2020, July). Learning representations in Bayesian Confidence Propagation neural networks. In *2020 International Joint Conference on Neural Networks (IJCNN)* (pp. 1-7). IEEE.

[6] Diehl, Peter U., and Matthew Cook. "Unsupervised learning of digit recognition using spike-timing-dependent plasticity." *Frontiers in computational neuroscience* 9 (2015): 99.

[7] Kheradpisheh, Saeed Reza, et al. "STDP-based spiking deep convolutional neural networks for object recognition." *Neural Networks* 99 (2018): 56-67.

[8] Mozafari, Milad, et al. "Bio-inspired digit recognition using reward-modulated spike-timing-dependent plasticity in deep convolutional networks." *Pattern recognition* 94 (2019): 87-95.

[9] Hao, Yunzhe, et al. "A biologically plausible supervised learning method for spiking neural networks using the symmetric STDP rule." *Neural Networks* 121 (2020): 387-395.

[10] Dong, Yiting, et al. "An unsupervised spiking neural network inspired by biologically plausible learning rules and connections." *arXiv preprint arXiv:2207.02727* (2022).